\documentclass{article}
\usepackage{spconf,amsmath,graphicx}

\usepackage{algorithmic}
\usepackage{algorithm}

\usepackage{algorithmic}
\usepackage{algorithm}
\usepackage{amsmath}
\usepackage{graphicx}
\usepackage{caption}
\usepackage{amsmath}
\usepackage{subfigure}

\title{Accelerate 3D Object Processing via Spectral Layout}
\name{Yongyu Wang}
\address{University of Southern California}

\begin{document}
%
\maketitle
\begin{abstract}

3D image processing is an important problem in computer vision and pattern recognition fields. Compared with 2D image processing, its computation difficulty and cost are much higher due to the extra dimension. To fundamentally address this problem, we propose to embed the essential information in a 3D object into 2D space via spectral layout. Specifically, we construct a 3D adjacency graph to capture spatial structure of the 3D voxel grid. Then we calculate the eigenvectors corresponding to the second and third smallest eigenvalues of its graph Laplacian and perform spectral layout to map each voxel into a pixel in 2D Cartesian coordinate plane. The proposed method can achieve high quality 2D representations for 3D objects, which enables to use 2D-based methods to process 3D objects. The experimental results demonstrate the effectiveness and efficiency of our method.

\end{abstract}
\begin{keywords}

\end{keywords}
\section{Introduction}
\label{sec:intro}

3D object processing is an important problem in computer vision, pattern recognition and image processing fields. Processing 3D object is a challenging task due to the following reasons: 1) 3D object has more complex representation compare to 2D image. 3D object usually represented by volumes or point clouds. In contrast, 2D image can be represented simply by pixel grid \cite{shen2019survey}. 2) Typical size of 3D representation is usually very large \cite{yang20193d}. So, the difficulty of processing 3D object is much higher compared to 2D object.

In recent years, substantial effort has been devoted to effectively process 3D object, and numerous methods have been developed. \cite{su2015multi,hegde2016fusionnet,qi2016volumetric} proposed to use multi-view representation of point cloud to handle 3D object classification tasks. \cite{chen2017multi} proposed to encode 3D point cloud using multi-view feature maps. \cite{engelcke2017vote3deep,guo20153d} proposed to use feature-based deep convolutional neural networks for shape classification. \cite{zhang2019three} proposed a regularization based pruning method for 3D CNN acceleration; \cite{su2015multi} proposed to use varying camera extrinsics to extract image features for different views. \cite{howard2017mobilenets} proposed to build light weight deep neural networks by using depth-wise separable convolutions. However, none of these methods achieve one-to-one mapping between 3D object and 2D images, so that the computational complexity of processing 3D object can be alleviated but not fundamentally solved. 

In this work, we propose to generate one-to-one mapping between 3D object and 2D image via spectral layout. By mapping 3D object into 2D image, we can use 2D-based methods to do 3D tasks. Experimental results demonstrate the effectiveness of our method.

\section{Preliminaries}\label{sect:Preliminaries}

\subsection{k-Nearest Neighbor Graph }
The underlying manifold structure has proven to be useful for improving the performance of image processing and computer vision tasks such as shape retrieval and image clustering \cite{premachandran2013consensus,yan2020self}. Among various manifold representation methods, $k$-nearest neighbor (k-NN) graph is most widely used due to its superior capability of capturing the local manifold \cite{roweis2000nonlinear}. In $k$-nearest neighbor graph, each node is connected to its $k$ nearest neighbors. The algorithm is shown in Algorithm \ref{alg:knn}.

\begin{algorithm}[!h]
\small { \caption{$k$-NN graph construction} \label{alg:knn}
\textbf{Input:} Data samples $x_1,...,x_n \in {R}^{d}$, neighborhood size $k$.\\
\textbf{Output:} Graph $G$.\\
    \begin{algorithmic}[1]

     \FOR {each data sample $x_i$}
     	\STATE{Compute distances between $x_i$ and all the other data samples};
        \STATE{Sort the computed distances};
        \STATE{Connect $x_i$ with its $k$ nearest data samples};
     \ENDFOR
     
    \end{algorithmic}
    }
\end{algorithm}

\subsection{Spectral Graph Theory}

Consider a graph $G=(V,E,w)$, where $V$ is its vertex set and $E$ is its edge set, $w$ denotes the weight function that assigns positive weights to all of its edges. The elements of its graph Laplacian matrix $L_G$ are given by:

\begin{equation}\label{formula_laplacian}
L_G(p,q)=\begin{cases}
-w(p,q) & \text{ if } (p,q)\in E \\
\sum\limits_{(p,t)\in E}w(p,t) & \text{ if } (p=q) \\
0 & \text{otherwise }.
\end{cases}
\end{equation}

Spectral graph theory uses the the eigenvalues and eigenvectors of graph Laplacian matrix to study the properties and structure of a graph. \cite{chung1997spectral} have shown that second smallest eigenvalue and its corresponding eigenvector can provide a good approximation to the optimal cut. Spectral clustering \cite{von2007tutorial} makes use of the bottom eigenvectors of the adjacency matrix of the data to detects clusters.

\subsection{Convolutional Neural Networks}
\begin{figure}[!h]
\centering\includegraphics[scale=0.25]{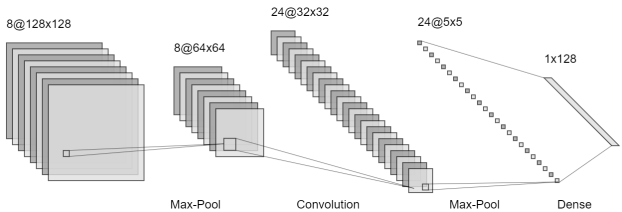}
\caption{architecture of LeNet-5 CNN.\protect\label{fig:Lenet5.eps}}
\end{figure}

Convolutional Neural Networks (CNNs) have became the most popular tool in machine learning fields. It has powerful ability in the context of learning features and image descriptors \cite{yan2020self}. The architectures of 2D CNNs are simple, consisting of an input layer, hidden layers and an output layer, as shown in in Fig. \ref{fig:Lenet5.eps}.

While it is natural to generalize 2D CNN to 3D, 3D CNNs often have more complex architecture with significantly more feature maps and parameters, making the training process very difficult and prone to over-fitting \cite{tran2015learning}.

 \section{Methods}
 \subsection{3D adjacency graph construction}
\begin{figure}[!h]
\centering\includegraphics[scale=0.2]{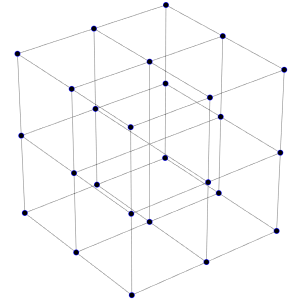}
\caption{3D adjacency graph.\protect\label{fig:3dknn}}
\end{figure}

In 2D space, graphs are often used to capture the relationship of entities to analyze the underlying structure. The most common way is to depict each entity as a node and use node-link diagram to connect closely related entities. In this work, we extend this representation to 3D space by connecting each voxel in 3D voxel grid with its adjacent voxels to form a 3D graph, as shown in Fig. \ref{fig:3dknn}, where the nodes in the graph represent the voxels and the dash lines represent edges between them.
Then we calculate the Laplacian matrix $L_G$ corresponding to this 3D adjacency graph.

\subsection{Spectral Layout}
Spectral layout aims to place nodes in high-dimensional space in a two dimensional plane using eigenvectors of a matrix \cite{koren2005drawing}.

\begin{figure}[!h] \centering 
\subfigure[3D adjacency graph]{ \includegraphics[width=2.1in,height=2.1in]{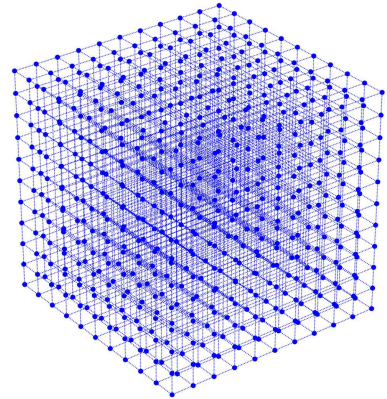} } \subfigure[2D grid]{ \includegraphics[width=2.5in,height=1.8in]{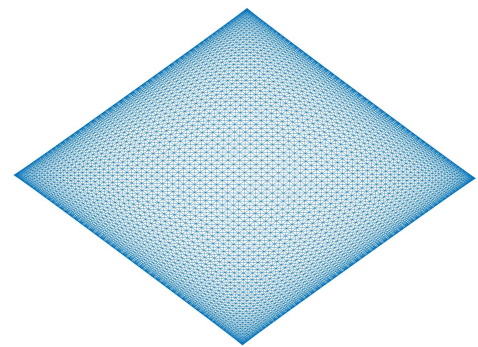} }  
\caption{Mapping from 3D space to 2D plance} \label{fig:accscaling} 
\end{figure}

Using the eigenvectors corresponding to the bottom eigenvalues has proven to significantly improve performance of recognizing complex patterns \cite{von2007tutorial}. In this paper, we first calculate the two eigenvectors corresponding to the second and third smallest eigenvalues of Laplacian matrix of the 3D adjacency graph. Then, we use the entries of these two eigenvectors as the 2D Cartesian coordinates for locating each voxel in a 2D Cartesian coordinate plane, as shown in Fig. \ref{fig:accscaling}. 

\subsection{Aggregate voxel values to form pixel intensities}

One of the novelties of our method is that the size of embedded 2D image can be set arbitrarily. On the 2D Cartesian coordinate plane, the start points and end points of the X-axis and Y-axis are set to the minimum values and maximum values of the entries of eigenvectors corresponding to the second and third smallest eigenvalues, respectively. Then, we divide the interval in subintervals of equal length. The number of subintervals is equal to the desired dimension of the embedded 2D image. Each square on the Cartesian coordinate plane is used as a pixel of the embedded 2D image. If the dimension is small, multiple voxels will be mapped into the same pixel. We simply sum their voxel values to form this pixel's intensity. The complete algorithm flow has been shown in Algorithm \ref{alg:kaspnew}.

\begin{algorithm}[!h]
\small { \caption{Spectral layout-based 3D object processing} \label{alg:kaspnew}
\textbf{Input:} A 3D object and desired 2D image dimension ${dim}$.\\
\textbf{Output:} A 2D image.\\

\begin{algorithmic}[1]
    \STATE Construct a adjacency graph $G$ to represent the spatial structure of a 3D voxel grid ; \\
    
    \STATE Compute the Laplacian matrix $L_G$ corresponding to graph $G$;\\
    \STATE Calculate the eigenvectors $u_2$ and $u_3$ of $L_G$; \\

    \STATE Perform spectral layout using $u_2$ and $u_3$  to map the 3D grid in to 2D plane ;\\
    \STATE Partition the 2D Cartesian coordinate plane into $dim\times dim$ squares.\\
   
    \STATE Map each voxel into a square on 2D Cartesian coordinate plane based on its entries of $u_2$ and $u_3$.\\
    \STATE Use each square on 2D Cartesian coordinate plane as a pixel of 2D image and sum the voxels in it to form its pixel intensity.\\

\end{algorithmic}
}
\end{algorithm}

https://modelnet.cs.princeton.edu/

\section{Experiments}

\subsection{Classification Result}
In this paper, we sample 5 categories ( cup, bowl, laptop, lamp, and stool) of the ModelNet40 dataset to perform classification. ModelNet40 is available on Princeton ModelNet data set \footnote{https://modelnet.cs.princeton.edu/}. 

We set the 2D image dimension to $144\times 144$ and use the simplest 2D CNN with one convolutional layer, one pooling layer and two dense layers. By training only 2 epochs, we got 91\% classification accuracy.

\subsection{Embedding Result}
We show the embedded 2D images ( $224\times 224$ dimension) of some categories to demonstrate the mapping quality.

\begin{figure}[!h]
\centering\includegraphics[scale=0.25]{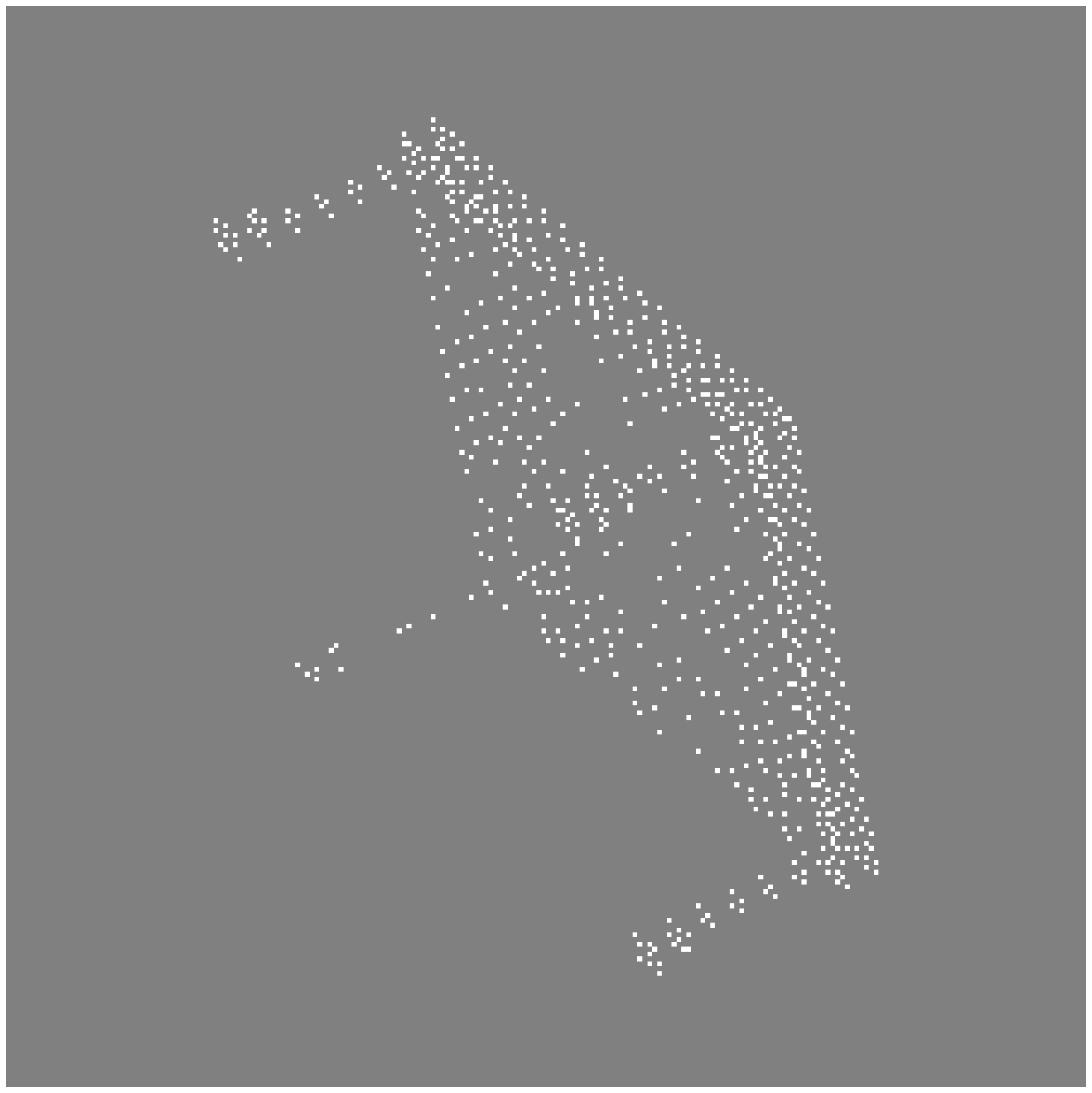}
\caption{embedded 2D table.\protect\label{fig:table1.eps}}
\end{figure}

\begin{figure}[!h]
\centering\includegraphics[scale=0.25]{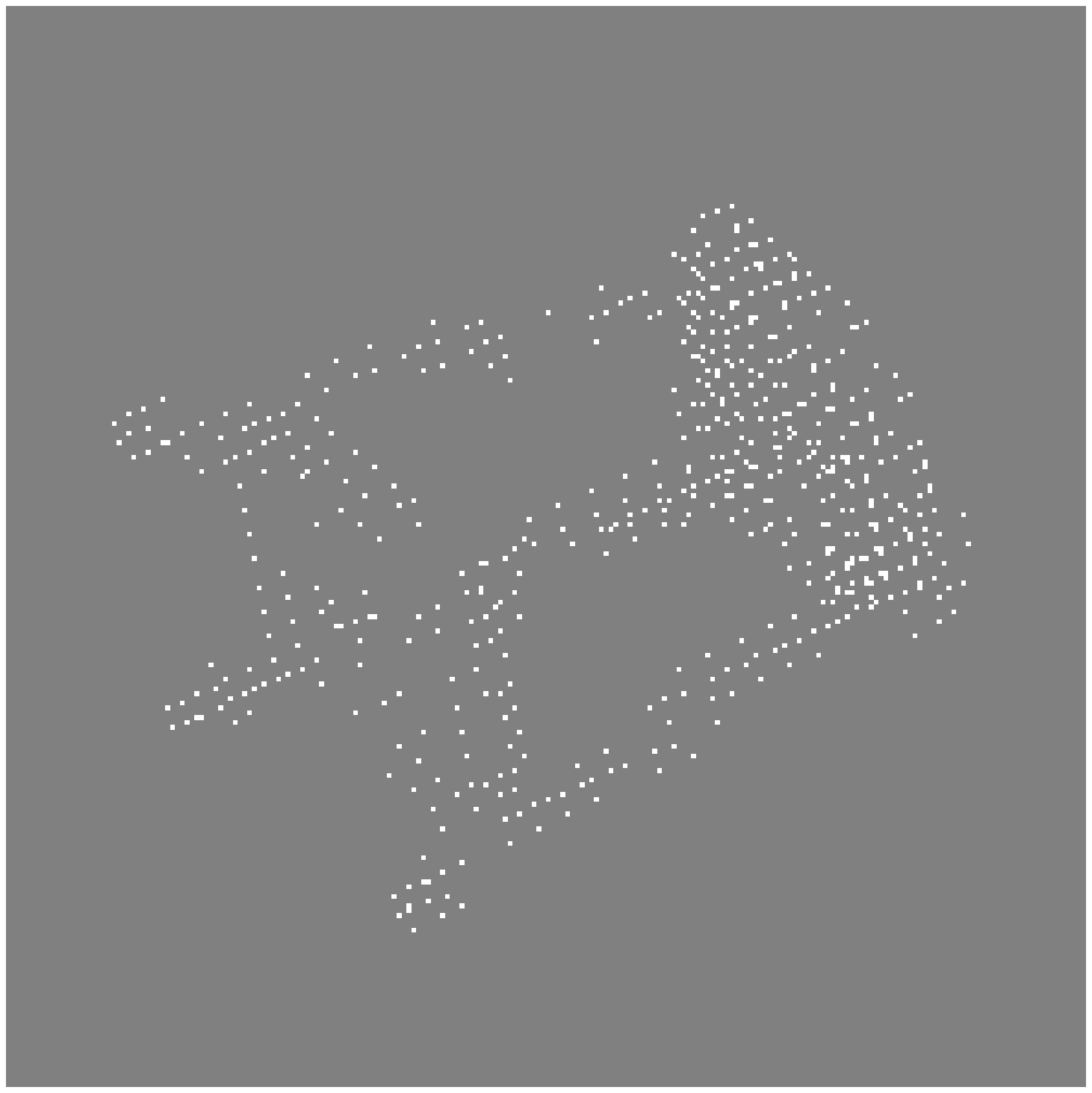}
\caption{embedded 2D stool.\protect\label{fig:stool1.eps}}
\end{figure}

\begin{figure}[!h]
\centering\includegraphics[scale=0.25]{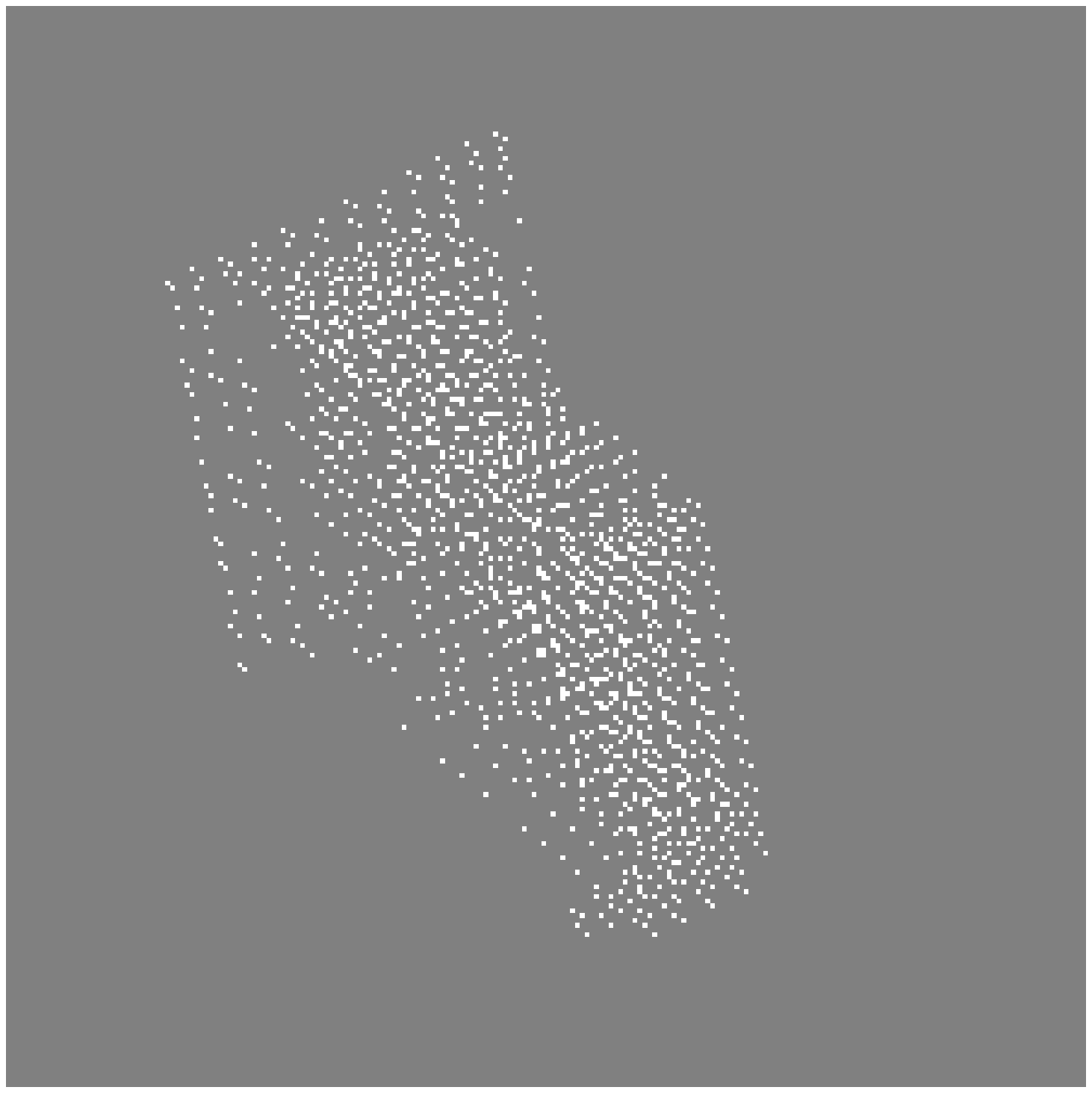}
\caption{embedded 2D bed.\protect\label{fig:bed1.eps}}
\end{figure}

\begin{figure}[!h]
\centering\includegraphics[scale=0.25]{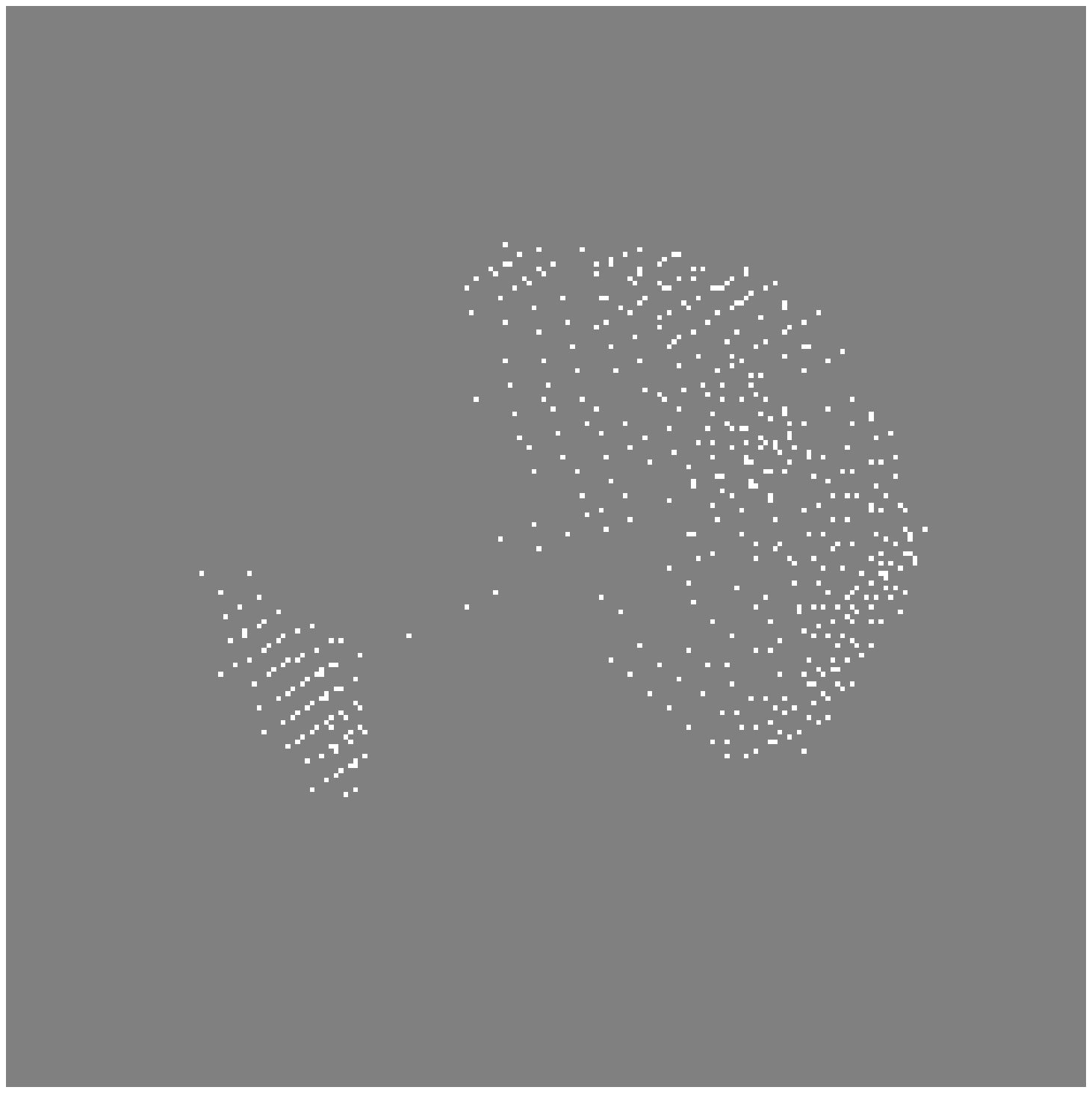}
\caption{embedded 2D lamp.\protect\label{fig:lamp1.eps}}
\end{figure}

\begin{figure}[!h]
\centering\includegraphics[scale=0.25]{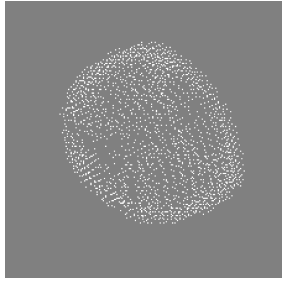}
\caption{embedded 2D bowl.\protect\label{fig:bowl1.eps}}
\end{figure}

\begin{figure}[!h]
\centering\includegraphics[scale=0.25]{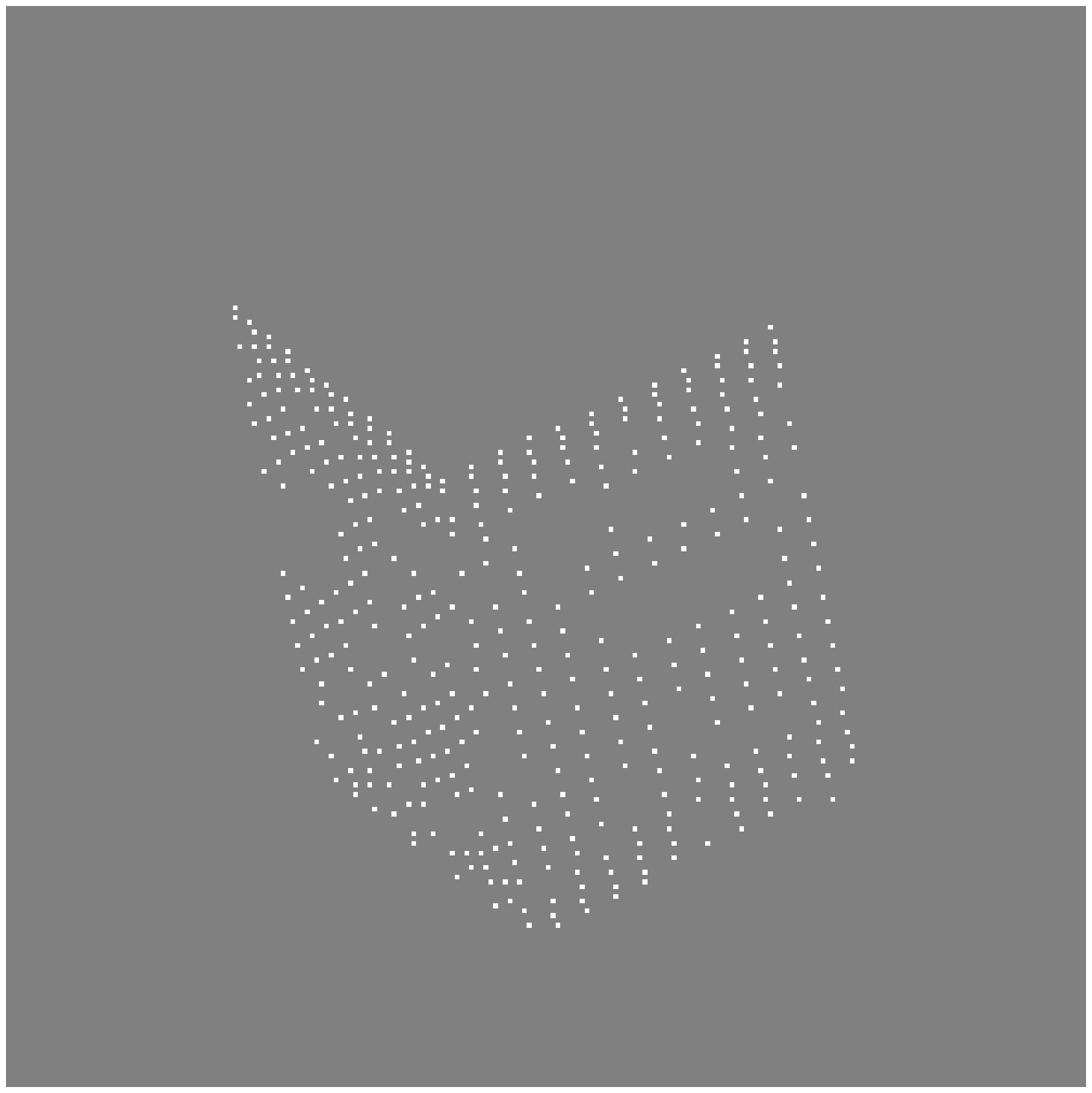}
\caption{embedded 2D laptop.\protect\label{fig:laptop1.eps}}
\end{figure}

\begin{figure}[!h]
\centering\includegraphics[scale=0.25]{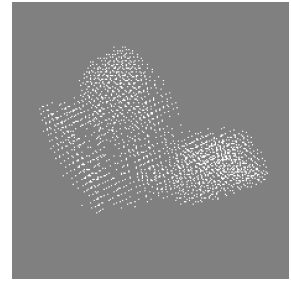}
\caption{embedded 2D toilet.\protect\label{fig:toilet1.eps}}
\end{figure}

\begin{figure}[!h]
\centering\includegraphics[scale=0.25]{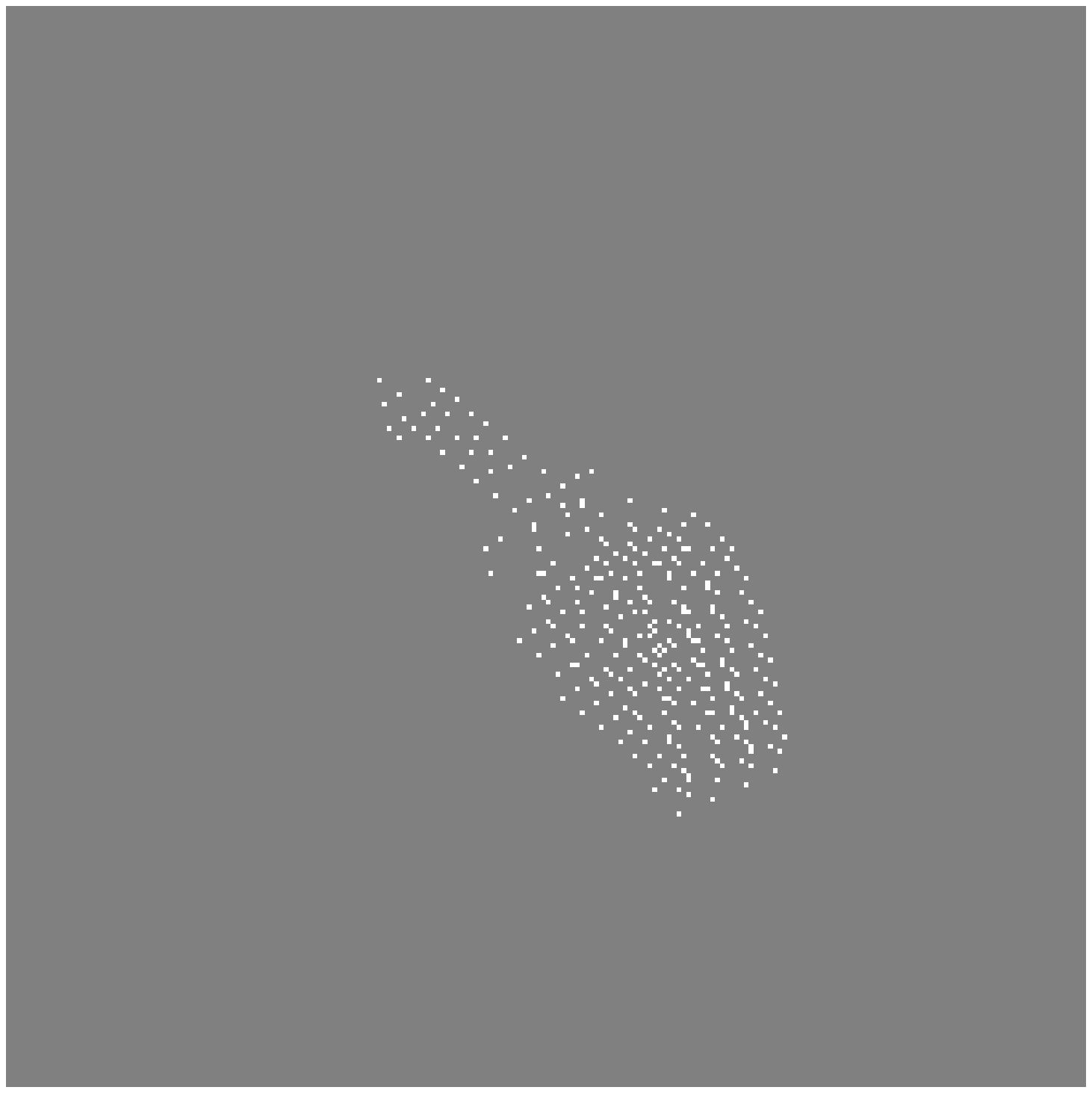}
\caption{embedded 2D guitar.\protect\label{fig:guitar1.eps}}
\end{figure}

\section{Conclusion}

In this paper, we use spectral layout the provide very high quality embedding of 3D objects. Our method enables us to use very simple 2D CNNs to process 3D objects with guaranteed solution quality. 

\section{Acknowledgments}
Some explorations of this work were made during 2017 Fall and 2018 Spring semesters when Yongyu Wang was a student at Michigan Technological University. The authors would like to thank Zhuo Feng for his helpful discussions during that period.

\vfill\pagebreak

\label{sec:refs}



\bibliographystyle{abbrv}
{
\bibliography{kdd17}  
}

\end{document}